# A Behavior Analysis-Based Game Bot Detection Approach Considering Various Play Styles

Yeounoh Chung, Chang-yong Park, Noo-ri Kim, Hana Cho, Taebok Yoon, Hunjoo Lee and Jee-Hyong Lee.

*An approach for game bot detection in MMORPGs is proposed based on the analysis of game playing behavior. Since MMORPGs are large scale games, users can play in various ways. This variety in playing behavior makes it hard to detect game bots based on play behaviors. In order to cope with this problem, the proposed approach observes game playing behaviors of users and groups them by their behavioral similarities. Then, it develops a local bot detection model for each player group. Since the locally optimized models can more accurately detect game bots within each player group, the combination of those models brings about overall improvement. For a practical purpose of reducing the workloads of the game servers in service, the game data is collected at a low resolution in time. Behavioral features are selected and developed to accurately detect game bots with the low resolution data, considering common aspects of MMORPG playing. Through the experiment with the real data from a game currently in service, it is shown that the proposed local model approach yields more accurate results.*

I. Introduction

Game bots are automated programs that perform repetitive tasks on behalf of human players. Since game bots can perform tedious tasks without break, those are usually used to obtain

Manuscript received March xx, 2013; revised March xx, 2013; accepted July xx, 2013.

This work was supported by Ministry of Culture, Sports and Tourism (MCST) and Korea Creative Content Agency (KOCCA) in the Culture Technology (CT) and Research Development Program 2012.

Yeounoh Chung (+82 31 290 7987, yeounohster@gmail.com), Chang-yong Park (codep@skku.edu), Noo-ri Kim (pd99j@skku.edu), Hana Cho (cho2405@skku.edu), and Jee-Hyong Lee (corresponding author, john@skku.edu) are with the School of Information and Communication Engineering at Sungkyunkwan University, Suwon, Rep. of Korea.

Taebok Yoon (tbyoon@seoil.ac.kr) is with the Department of Computer Software at Seoil University, Seoul, Rep. of Korea.

Hunjoo Lee (hjoo@etri.re.kr) is with the Contents Research Division, ETRI, Daejeon, Rep. of Korea.

http://dx.doi.org/10.4218/etrij.13.0112.0061

unfair advantages over honest human players. An unruly swarm of game bots depletes game contents and resources, so honest players may feel discouraged and lose interest, and eventually retire from the game. Game bots are potentially serious threats to gaming businesses, and detecting game bots is of a great importance to game publishers.

There have been many efforts on the game publisher side to prevent the use of game bots. However, it is hard to detect game bots because they simulate legitimate human game playing behavior. Various strategies have been proposed for game bot detection, such as repeated Turing test, network traffic analysis, bot scanning, but those are known to have drawbacks: interfering with game play or being easily evaded by game bots [1], [4]-[6].

A promising alternative to the traditional bot detection approaches is behavior analysis of game playing [2]. To this end, researchers analyzed behavioral patterns of players. However, the previous work tried to discriminate human players and game bots with one global model. Players have various gaming styles; some users are battle-oriented, some are item collection-oriented, and some are quest-oriented. Since such players exhibit different and distinct behavioral patterns, behavior-based bot detection should use a different set of bot detection rules for each type of player groups. However, previous researches did not consider the differences in behavioral patterns of player groups. Instead, a global bot detection model was used for the different player groups.

Another limitation of the previous work is that they only examines one or two characteristic behaviors for bot detection, rather than whole aspects of behavior [12], [16], [22]. Observing one or two types of behaviors can still accurately detect game bots; however, because distinguishing patterns for bot detection vary greatly from game to game, the previous approaches can be highly game dependent. For instance, if detecting game bots based on party play (party duration) [3], then the detection mechanism works well only with game bots in parties, but does not effectively work for players who do not party nor share their collections. Thus, it is important for a generic bot detection methodology to consider whole aspects of behavior in order to cover game bots with different gaming styles.

Another concern for behavior-based bot detection approaches is the game server workload for collecting game play data of all the players. In general, behavior-based game bot detection approaches require game play data in high resolution in time; however, collecting such high resolution data for every player can overload the game servers in service. Although, it is important to minimize the workload of the game servers in collecting game play data, most of the previous behavior-based bot detection approaches did not address this concern [12], [17], [20].

In this paper, we propose a novel behavior analysis-based bot detection methodology. We group players by their behavioral similarities and detect game bots within each group, using a set of customized bot detection rules for each group of players. We overcome the limitation of using a global bot detection model for different player groups by using local models for player groups, grouped by behavioral similarities.

Additionally, we examine the behavioral features that can be extracted from the low resolution game data to reduce the game server workloads. The features are common in MMORPGs and reflect the whole aspects of playing behavior. Features reflecting the whole aspects of playing behavior are necessary to effectively detect game bots with low resolution data, as well as to reduce the game-dependency of the proposed method.

We evaluate the proposed method using a large dataset from an MMORPG currently serviced by a Korean game publisher. Our experimental results confirm the importance of considering various aspects of game behavior and different gaming styles to detect game bots.

The rest of this paper is organized as follows. Section 2 reviews related work in game bot detection. Section 3 describes the features used for our experiment and the proposed bot detection framework. Section 4 presents and explains the experimental results. Section 5 summarizes and concludes the work.

## II. Related Work

It is difficult to detect game bots because they simulate legitimate human game playing behavior. In an effort to restrain game bots in MMORPGs, various strategies have been proposed, such as repeated Turing test, network traffic analysis, bot scanning and behavior analysis.

The repeated Turing test, such as CAPTCHA (Completely Automated Public Turing test to tell Computers and Humans Apart) provides a good way to detect game bots [4], [5]. The basic idea of the Turing test is to ask a question that is easy for humans to answer, but very difficult for computers. Based on the answer from players, one judges if the respondent is human or not. The CAPTCHA authentication is widely used in MMORPGs for bot detection [6]-[11]. However, Q&A based approaches interfere with normal game play, and some advanced game bots provide an evasion function for CAPTCHA authentication [13], [15].

Chen et al. [12] traced game data packets for bot detection. They showed that there were differences in network traffic patterns generated by humans and game bots. Bot detection measures based on network traffic analysis are less obtrusive to human players than Turing test or anti-bot software [14], but they can cause network overload and lags in game play [22].

Client-side bot scanning is one of the traditional anti-bot methodologies. Anti-bot software for bot scanning is installed on client computers and examines event sequences and memories. This can cause inconveniences for players, such as collisions with operating systems [22]. In many cases, such bot detection mechanism can be evaded by running games in guest mode on an administrator account [20].

A promising alternative to the above approaches is bot detection based on player behavior analysis. To this end, researchers analyzed various behavioral patterns of game players using machine learning or statistical techniques.

Some researchers focused on movement pattern of game play behavior [16]-[19]. This exploits the fact that bots move along prescribed paths, thus show regularities in their movement patterns, while human players follow more complex and random movement paths. However, bot writers can easily introduce some irregularities to bots' movement patterns [20].

Other features that are preferred for bot detection are attack sequences [12], [21] and social sequences [18], [23], [24], because bots usually have regular patterns in those features. However, they are also limited in that such features may not be common or relevant in game bots, depending on games or gaming purposes.

One recent study analyzed party play behavior to detect game bots. The authors showed that game bots tend to stay in party much longer (almost indefinitely) than human players [22]. This approach also has limitations; their bot detection mechanism solely depended on party play behavior, which may not work for game bots playing individually.

A limitation of the approaches mentioned above is that they considered a single aspect of player behavior. The features of a particular behavioral aspect may be very effective in some games but may not work effectively in some others to detect bots. Furthermore, even if the features of the particular behavioral aspect are effective, they may not be practically applicable if collecting game data for the features overloads the game servers in service. Hence, more generic and practically applicable features need to be observed.

Table 1. Description of behavioral features

| Feature | Description |
|---|---|
| Hunting | Accumulated number of hunted NPCs for a sampling interval |
| Attack | Accumulated attack count for a sampling interval |
| Hit | Accumulated hit (successful attack) count for a sampling interval |
| Defense | Accumulated defense count for five minutes |
| Avoidance | Accumulated avoidance (successful defense) count for a sampling interval |
| Recovery | Accumulated number of healing portion usages for a sampling interval |
| Item | Number of items at the time of sampling |
| Collection | Accumulated number of collected items for a sampling interval |
| Drop | Accumulated number of dropped items for a sampling interval |
| X | Coordinate X at the time of sampling |
| Y | Coordinate Y at the time of sampling |
| Portal | Accumulated number of portal item usage count for a sampling interval |

Table 2. Developed feature set

| Feature | Formula |
|---|---|
| Combat ability Ⅰ | $\frac{Hunting}{Attack} + \frac{Avoidance}{Defence}$ |
| Combat ability Ⅱ | $\frac{Recovery}{Hunting}$ |
| Combat ability Ⅲ | $\frac{Hunting}{\sqrt{|X - pre\_X|^2 + |Y - pre\_Y|^2}}$ |
| Collect pattern | $\frac{(Collection + Drop)}{2 \times Item}$ |
| Move pattern | $\sqrt{|X - pre\_X|^2 + |Y - pre\_Y|^2}$ |

Another limitation is that they did not consider the fact that players have quite different behavior patterns. The previous work tried to discriminate human and bot players with one global model. Since there are various game playing styles, it is difficult for a single model to effectively handle all the players. The overall performance of a global model may be low and some specific types of players may not be correctly classified.

## III. Proposed Method

In this section, we propose a bot detection approach based on player behavior analysis. The proposed bot detection approach works in two steps. First, players are clustered by their behavioral similarities. Second, a local model with customized bot detection rules is generated for each cluster. To describe the proposed methodology, we first explain the features selected and developed for low resolution data, and then bot detection model construction considering different types of players.

### 1. Feature Selection

In general, behavior-based bot detection methods can yield accurate results because such methods exploit key behavioral patterns that are different in humans and game bots. However, collecting and monitoring game behavior data for every player can be very expensive. Therefore, behavioral features that are effective and less burdening to the systems need to be carefully chosen and developed.

In order to reduce the game server workloads in data collection, we select simple behavioral features that are extractable from the game data collected at a low sampling rate. Some of the simple behavioral features are attack counts, hit counts and item counts, which are either accumulated values during the sampling interval or snapshot values at the end of the each sampling interval.

Since we assume a low resolution of the collected data, some of the informative behavioral features used in the previous researches are not used in this study. For instance, repeated moving paths and attack sequences are good features for behavior-based bot detection [17], [20], but those features require the game data sampled at a fine resolution.

The features are selected considering the three categories of game behavior: *Battle*, *Collect* and *Move*. These categories are common in MMORPGs and thus they can represent important aspects of game behavior and players with various play styles. A general bot detection approach should consider various aspects of playing behavior, because distinguishing patterns for bot detection vary greatly from game to game. Furthermore, the data sampling resolution is very low, so it needs to consider various aspects of game playing behavior.

For *Battle* category, we choose *Hunting*, *Attack*, *Hit*, *Defense*, *Avoidance*, and *Recovery*; for *Collect* category, *Item*, *Collection* and *Drop*; for *Move* category *X*, *Y* and *Portal*, which are very common in most MMORPGs. The features presented in Table 1 are designed based on those behaviors. So, those features are generally applicable and can be evaluated from game playing behaviors in most MMORPGs.

As mentioned, the selected features are just accumulated or snapshot values. For example, the attack feature, *Attack*, indicates how many times a player attacked within a sampling interval on average, but the feature does not convey any information on how effectively the player attacked. In order to grasp deeper meanings of player actions, we develop new features based on the selected feature set: *Combat ability I, II,*

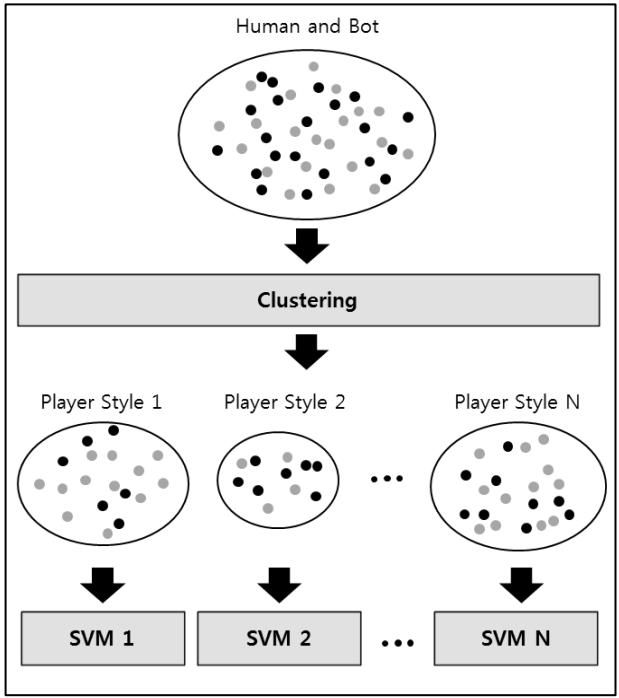

**Figure 1. Bot detection considering different game play styles**

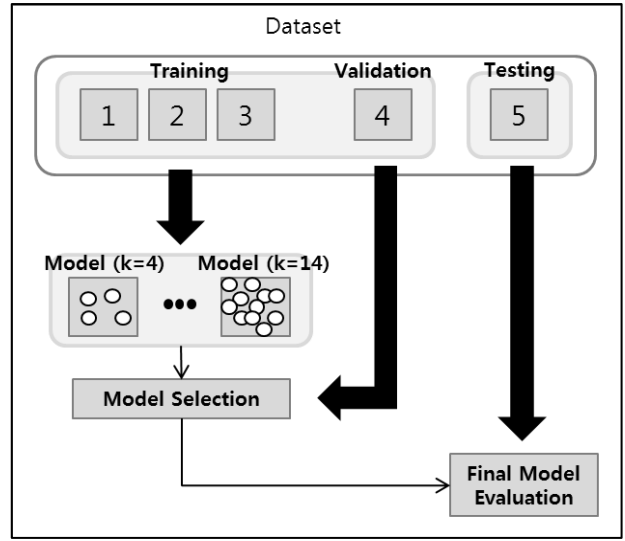

**Figure 2. Process for finding proper number of clusters, $k$**

*III*, *Collect pattern* and *Move pattern*. The formal definitions of the developed features are shown in Table 2.

These features are developed to measure the effectiveness and efficiency of player behavior. *Combat ability I* measures how effectively a player attacks and defends, by taking the attack success rate and defense success rate. *Combat ability II* measures how health-efficiently a character combats based on the number of healing portion usages per hunting (killing a monster). Using a lesser number of healing portions for every hunt indicates that the player manages his or her health efficiently in combat. *Combat ability III* measures how space-efficiently a player battles based on the number of hunted monsters and traveled distance. Hunting more monsters in a shorter traveled distance indicates that the player efficiently hunted monsters.

*Collect pattern* measures how efficiently a player uses his or her item slots. Emptying full item slots and filling empty item slots represent efficient use of item slots. In addition, *Move pattern* is evaluated to observe the range of movement within a fixed time period. This feature does not measure the effectiveness or efficiency of player behavior, but there can be a correlation between the range of movements and play styles. For example, quest-oriented players would travel more distance than players with other gaming purposes

Both the selected and the developed features are standardized with Z-scores in order to equally weight each behavioral feature. This step is of particular importance, because features have different value ranges.

### 2. Player Grouping and Bot Detection Model Generation

Game players have different gaming purposes and behavioral patterns, which results in various game play styles. So, it is necessary in behavior analysis based bot detection to consider various game play styles and use a different set of bot detection rules or a local model for each play style.

In this paper, we propose a bot detection methodology, concerning various game play styles in MMORPGs. We first examine three aspects (*Battle*, *Collect* and *Move*) of player behavior that are relevant to common MMORPG play, and group players by their behavioral similarities based on player behavior. Players who have similar gaming purposes and preferences may also exhibit similar behavioral patterns. Hence, we can group players by their game play styles using their behavioral similarities.

Figure 1 gives an overview of the proposed methodology. We use a *k*-means algorithm for grouping players by their behavioral similarities. The *k*-means algorithm is a general clustering technique for grouping data into *k* clusters, for a given *k*. Given player data with behavioral features, the *k*-means algorithm groups players who have similar behavior patterns, by maximizing intra-cluster behavioral similarities, and minimizing inter-cluster behavioral similarities. So, we can obtain *k* player groups, in which players have similar game playing patterns.

For each cluster of similar players, an SVM bot detection model is generated. The local SVM model is trained with human and bot player data from the associated cluster. SVMs are a well-known machine learning technique used in binary classification. SVMs require only a few parameters for model tuning, and the method is well-known for its low generalization error [25]. Since the proposed method develops local models for each group, local models can be best customized to detect game bots in each player group. Such local bot detection models would more accurately and effectively detect game bots within each player group. The combination of these models can outperform an approach based on a single global model.

If an unknown player is given, the player is classified as a human player or a game bot, using the generated *k* clusters and the *k* SVM models. First, the most similar cluster to the player is identified. Euclidean distances between the player behavioral feature vector and cluster centroids are used as a similarity measure between players and clusters. If the player belongs to cluster *i*, then its local SVM model, $SVM_i$ is used. The behavioral feature vector of the player is inputted to $SVM_i$, and the player is classified as the output of $SVM_i$: a human player or a game bot.

Each player group may consist of both human and bot players with similar behaviors. Therefore, a local bot detection model for a particular player group can focus more on the behavioral differences between human and bot players, and less on the behavioral differences among different play styles, which may act as noises in using a global bot detection model. Hence, we can build local detection models which are well adapted to each group with higher accuracy.

However, if the number of groups is too small, players with less behavioral similarities can be clustered together. If the number of groups is too large, each cluster may contain too few players to build a bot detection model for the cluster. In such cases, the proposed method may be inaccurate. Hence, the number of clusters, *k*, should be carefully chosen. In this study, we tried 4 to 14 clusters and selected the number of clusters which showed the best performance using validation data.

Figure 2 describes the process for finding the proper number of clusters, *k*. The dataset is evenly and randomly divided into 5 folds. Then, 4 folds are used for model building, and the remaining 1 fold is used for evaluating the performance of the final model. At the model building step, we first build 11 models with 3 training folds. Each model has a different number of clusters from 4 to 14. Next, each model is evaluated using the validation data. The model that produces the maximum performance is selected as the final model. The final model is the output of the model building step. Finally, the performance of the final model is evaluated with the testing data, which is independent of the data for model building.

Table 3. Number of logs and characters

| Group | Number of logs | Number of players |
|---|---|---|
| Human player logs | 3,529,099 | 20,543 |
| Bot player logs | 12,968,264 | 10,282 |
| Summation | 16,497,563 | 30,825 |

Throughout the experiment, we used 5-fold cross-validation to reduce statistical variance.

## V. Experiment

### 1. Raw Data

We collected data from a popular Korean MMORPG game, titled "*Yul-Hyul-Gang-Ho Online*." The game has more than 200 million subscribers in total, with a maximum of 85,000 concurrent connections.

Most game publishers strictly prohibit the use of game bots, but "*Yul-Hyul-Gang-Ho Online*" allows players to purchase and use a game bot item, called "the Box of Black Soul." Since the game bots also perform repetitive tasks with given programmed rules as illegal games bots do, we may use the data of "*Yul-Hyul-Gang-Ho Online*" for developing a bot detection methodology.

Data was collected from a game server in service for a month. In order to minimize the workloads of the game server in service, the length of sampling intervals is set to five minutes, which is a long interval to obtain descriptive behavior data.

A brief summary of the collected behavior log data is shown in Table 3. In Table 3, the number of logs is the number of collected data log every five minutes. That is, 20,543 human players created 3,529,099 five minute data logs. There are twice more human players than game bot players, while there are 3.6 times more bot player logs than human player logs. In MMORPGs, there are generally more human players than bot players, but there are more bot players in "*Yul-Hyul-Gang-Ho Online*" because the use of game bots is allowed.

We extract bot player instances and human player instances at a one-to-one ratio. Since there are more bot players than human players, using the entire data instances would result in biased bot detection models. In order to generate more general unbiased bot detection models, we under-sample bot player instances.

Finally, we standardize features of data instances with Z-scores in order to equally weight each behavioral feature. This step is of particular importance, because features have different value ranges. Equation 1 shows the Z-score standardization.

$$f_i^z = \frac{f_i - m_f}{\sigma_f} \quad (1)$$

In the equation, $f_i$ is the value of feature *f* of instance *i*, $m_f$ and $\sigma_f$ are the average and the standard deviation of feature *f* of all players, respectively. $f_i^z$ is the Z-score of $f_i$.

### 2. Baseline and Evaluation Metrics

In order to compare the performance of the proposed method,

we use a global SVM bot detection model as the baseline, which does not consider game play styles. Most of the previous approaches were based on a single global model, so the model based on a single SVM is chosen as the baseline.

We used four evaluation metrics: *Accuracy*, *Precision*, *Recall*, and *F1*. *Accuracy* measures how many bots and humans are correctly identified. *Precision* measures how many of players detected as bots are really bots. *Recall* measures how many of the real game bots are detected. More formal definitions are as follows:

$$Accuracy = \frac{TP + TN}{TP + TN + FP + FN} \quad (2)$$

$$Precision = \frac{TP}{TP + FP} \quad (3)$$

$$Recall = \frac{TP}{TP + FN} \quad (4)$$

In the equations, *TP*, *TN*, *FP* and *FN* are the number of real bots identified as bots, the number of humans identified as humans, the number of humans identified as bots, and the number of bots identified as humans, respectively.

Usually, *Precision* and *Recall* tend to be inversely proportional to each other. If *Precision* is high, *Recall* is low; and vice versa. So, we observe one more measure, *F1*, which combines *Precision* and *Recall* as follows:

$$F1 = \frac{2 \cdot Precision \cdot Recall}{Precison + Recall} \quad (5)$$

This is the harmonic mean of *Precision* and *Recall*.

## 3. Bot Detection Results

In this experiment, we show that the proposed method outperforms the baseline. For the experiment, we use 6

**Table 5. Baseline result**

| | Feature | Accuracy | Precision | Recall | F1 |
|---|---|---|---|---|---|
| Whole aspect | $F_{17}$ | 94.27 | 94.69 | 93.90 | 94.29 |
| | $F_{12}$ | 84.72 | 81.42 | 87.17 | 84.20 |
| | $F_5$ | 88.07 | 95.17 | 83.33 | 88.86 |
| Single aspect | $F_B$ | 80.25 | 68.17 | 89.88 | 77.53 |
| | $F_M$ | 59.43 | 63.22 | 58.77 | 60.91 |
| | $F_C$ | 67.31 | 61.57 | 69.55 | 65.31 |

**Table 6. Proposed method result**

| | Feature | Accuracy | Precision | Recall | F1 |
|---|---|---|---|---|---|
| Whole aspect | $F_{17}$ | 96.20 | 95.95 | 96.42 | 96.18 |
| | $F_{12}$ | 91.50 | 91.68 | 91.35 | 91.51 |
| | $F_5$ | 89.61 | 93.62 | 86.66 | 90.01 |
| Single aspect | $F_B$ | 87.87 | 86.20 | 89.17 | 87.65 |
| | $F_M$ | 72.38 | 71.88 | 72.61 | 72.24 |
| | $F_C$ | 68.14 | 69.40 | 67.68 | 68.53 |

**Table 4. Feature sets used in the experiment**

| | Feature | Description |
|---|---|---|
| Whole aspect | $F_{12}$ | The 12 selected features in Section 3.1 |
| | $F_5$ | The 5 developed features in Section 3.1 |
| | $F_{17}$ | The 12 selected and the 5 developed features ($F_{12} \cup F_5$) |
| Single aspect | $F_B$ | Subset of $F_{12}$, containing only *Battle* aspect features |
| | $F_M$ | Subset of $F_{12}$, containing only *Move* aspect features |
| | $F_C$ | Subset of $F_{12}$, containing only *Collect* aspect features |

different feature sets for two reasons. First, it is to verify that the superiority of the proposed method over the baseline is independent of the feature sets. Second, it is to verify that the selected and developed feature sets are effective over single aspect feature sets.

The 6 feature sets used in the experiment are summarized in Table 4. $F_{12}$ is the 12 selected features in Section 3.1. $F_5$ is the 5 developed features in Section 3.1. $F_B$, $F_C$ and $F_M$ are subsets of $F_{12}$, categorized as battle, collect, and move feature groups in Section 3.1, respectively. Lastly, $F_{17}$ is the union of $F_{12}$ and $F_5$. $F_{17}$, $F_{12}$ and $F_5$ are whole aspect feature sets, and $F_B$, $F_C$ and $F_M$ are single aspect feature sets.

Table 5 shows the baseline results, and Table 6 shows the proposed method results. For any feature sets, the proposed approach gives superior results to the baseline. In *Accuracy*, the performance of the proposed method is higher than that of the baseline under any feature sets. In *Precision* and *Recall*, the baseline is a little higher in some cases. However, in *F1*, a measure considering both *Precision* and *Recall* together, the proposed method shows much better performance.

In the case of $F_{12}$, which does not include the features developed by experts, the proposed approach shows much better results than the baseline. The accuracy, precision, and recall are higher by 8.0%, 12.6% and 4.8% than those of the baseline, respectively. However, in the case of $F_{17}$ where $F_5$ is considered together with $F_{12}$, the performance gap is reduced. The proposed method shows about 2% better performance in the cases of $F_{17}$ and $F_5$. From this, we may conclude that well-developed features from expert knowledge are effective for bot detection.

The performances of the single aspect feature sets are lower than those of the whole aspect feature sets, which shows why we have to consider the whole aspects of behavior. Among the single aspect features, $F_B$ gives the most accurate results in either methodology. The performance is very close to that of $F_{12}$. This implies that the battle feature group may represent the most important aspect of player behavior in the game, "*Yul-Hyul-Gang-Ho Online*", in clustering and bot detection.

## 4. Play Style Analysis

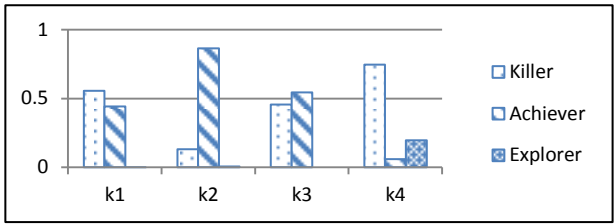

Figure 3. Player type ratio by $F_{17}$

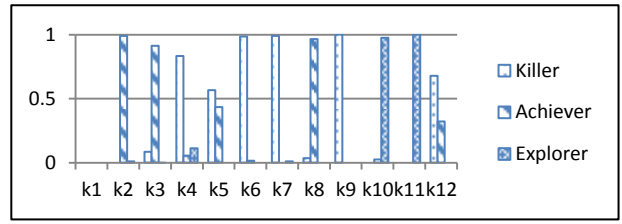

Figure 4. Player type ratio by $F_{12}$

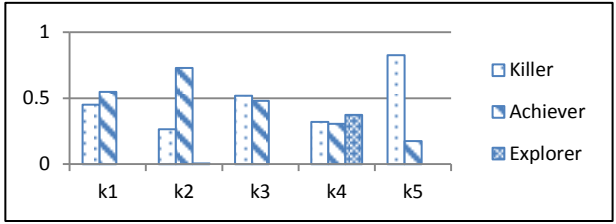

Figure 5. Player type ratio by $F_5$

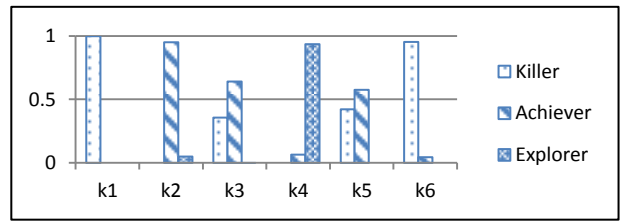

Figure 6. Player type ratio by $F_B$

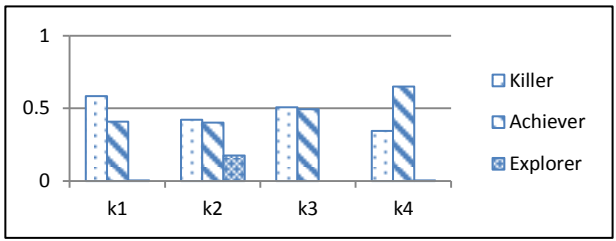

Figure 7. Player type ratio by $F_M$

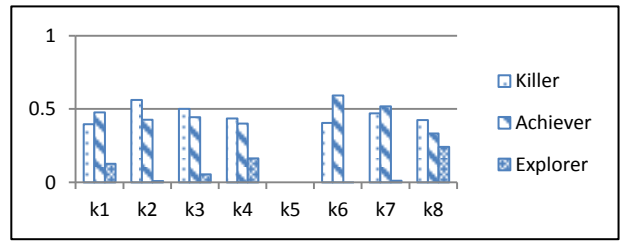

Figure 8. Player type ratio by $F_C$

In this subsection, we analyze how distinctively the proposed approach discriminates and groups similar play style users, and how accurately the proposed approach classifies human and bot players in each play style.

For this purpose, we first identify the play styles of all the players. According to the Bartle Test of Gamer Psychology [26], players in MMORPGs can be classified into four categories, based on their gaming styles, as shown in Table 7.

For the experiment, a group of five human experts played the game more than 100 hours and was given the same raw game playing data of the players. They were asked to identify the players who had strong characteristic of *Killer*, *Achiever* or *Explorer* with their own expert knowledge. Player data without any strong behavioral characteristic were simply ignored and later labeled as *Remainder*. Since there are no features for social behavior in the collected dataset, the experts could not identify *Socializer* player type.

Table 8 shows the player distribution of the three player types and the remainder. The majority of the players, both human players and game bots, are *Remainder*. A normal human player may not consistently play in a single play style, and thus may not be classified as one of the three player types. On the other hand, game bots continuously repeat the same programmed tasks, so can be more easily classified as one of the styles.

We verified how well the proposed method clusters players with different game play styles. For each feature set, we tried 4 to 14 clusters and selected the number of clusters with the best performance. Since the numbers of clusters for the best bot detection performance may be different depending on the feature sets, the numbers of clusters in the figures are different. Clustering with $F_{17}$, the proposed method yields the most accurate results when the number of clusters is 4. For each of those 4 clusters, Figure 3 shows the ratio of three types, *Killer*, *Achiever* and *Explorer* except *Remainder*, in each cluster. For example, the first cluster ($k_1$) consists of about 55% of *Killers*, 45% of *Achievers* and 0% of *Explorers*. The player clustering result by $F_{12}$, $F_5$, $F_B$, $F_M$, and $F_C$ are also shown in Figures 4~8. And the numbers of clusters for the best performance are 12, 5, 6, 4, and 8 for $F_{12}$, $F_5$, $F_B$, $F_M$, and $F_C$, respectively.

**Table 7. Four player types in MMORPG from the Bartle test of gamer psychology [26]**

| Player Type | Explanation |
|---|---|
| Killer | Players prefer fighting other players and engage in battles. |
| Achiever | Players who prefer to gain experiences (skill levels), items, and other game rewards. |
| Explorer | Players who find great joy in discovering areas, maps, and hidden places. |
| Socializer | A multitude of players who play games for the social aspect, rather than the actual game itself. |

**Table 8. Player type distribution**

|  | Killer | Achiever | Explorer | Remainder |
|---|---|---|---|---|
| **Human Players** | 777 | 1,266 | 224 | 6,186 |
| **Game Bots** | 2,296 | 1,895 | 11 | 4,251 |
| **Total** | 3,073 | 3,161 | 235 | 10,437 |

Table 9. Accuracy of play styles by baseline

| Feature | | Play Style | | | | Dev. |
|---|---|---|---|---|---|---|
| | | Killer | Achiever | Explorer | Remainder | |
| Whole aspect | $F_{17}$ | 92.42 | 95.57 | 95.32 | 94.39 | 1.59 |
| | $F_{12}$ | 86.59 | 87.88 | 94.89 | 82.98 | 2.54 |
| | $F_5$ | 88.48 | 82.70 | 96.17 | 89.39 | 3.63 |
| Single aspect | $F_B$ | 86.82 | 80.86 | 87.23 | 77.97 | 4.51 |
| | $F_M$ | 57.83 | 71.46 | 94.89 | 55.47 | 8.63 |
| | $F_C$ | 65.64 | 66.56 | 37.02 | 68.72 | 1.58 |

Table 10. Accuracy of play styles by proposed method

| Feature | | Play Style | | | | Dev. |
|---|---|---|---|---|---|---|
| | | Killer | Achiever | Explorer | Remainder | |
| Whole aspect | $F_{17}$ | 95.35 | 96.58 | 97.45 | 96.30 | 0.64 |
| | $F_{12}$ | 94.47 | 93.26 | 94.04 | 90.04 | 2.29 |
| | $F_5$ | 90.21 | 85.61 | 97.02 | 90.48 | 2.74 |
| Single aspect | $F_B$ | 90.17 | 89.12 | 86.38 | 86.84 | 1.70 |
| | $F_M$ | 77.25 | 74.18 | 95.30 | 69.89 | 3.70 |
| | $F_C$ | 69.83 | 68.11 | 27.66 | 68.55 | 0.89 |

The clustering result by $F_B$ is the best and the results by $F_{17}$, $F_{12}$ and $F_5$ are relatively good. On the other hand, the result by $F_M$ and $F_C$ are the worst. This result confirms that player types are hard to distinguish with only the move or collect feature set.

Figure 6 by the battle feature set shows an interesting result, in which players are very distinctively separated into different clusters by their play types. This is because the three types of players, *Killers*, *Achievers* and *Explorers*, have distinct values of battle features. *Killers* are battle-oriented players, so they have the most extreme values. On the other hand, *Explorers* are usually quest-oriented players, so they have the least extreme values. Even though *Achievers* engage in battles to gain experiences and collect items, hunting is not the sole purpose of *Achievers*. The battle related feature values for *Achievers* are less extreme than those of *Killers*, but more extreme than those of *Explorers*.

What is interesting is that play styles are the best separated by $F_B$ but human and bot players are not well separated, compared with the result by $F_{17}$, $F_{12}$ or $F_5$ in Tables 5 and 6. Since players are best clustered by $F_B$, $F_B$ might give the most accurate results for bot detection. However, the bot detection results are more accurate with $F_{17}$, $F_{12}$ or $F_5$. This also implies that various aspects of player behavior should be considered for not only grouping players by their styles, but also analyzing player behavior for bot detection.

Next, we analyze the classification performance of the baseline and the proposed method for each play style. Tables 9 and 10 show the bot detection accuracies for each game play styles. For example, for the *Killer* player group, the baseline has an accuracy of 92.42% with $F_{17}$, but the proposed method has an accuracy of 95.35% with $F_{17}$. The proposed approach shows better performance than the baseline in all the cases except (*Explorer*, $F_{12}$), (*Explorer*, $F_B$), (*Explorer*, $F_C$) and (*Remainder*, $F_C$). Since the *Explorer* player group is only 1.4% of the total player population, the result may not be meaningful in a statistical point of view, and it does not have much effect on the overall performance.

In addition to superiority in the overall accuracies, the proposed method is more stable. In Tables 9 and 10, the column *Dev.* shows the standard deviation of accuracies for *Killer*, *Achiever* and *Remainder*. Since the *Explorer* player group is 1.4% of the total player population, as mentioned, it is ignored. For example, in the case of $F_{17}$, the standard deviation of the baseline for *Killer*, *Achiever* and *Remainder* is 1.59 but that of the proposed method is 0.64. Under any feature sets, the proposed method shows smaller variances than the baseline, which means that the proposed method more stably produced better results. It also shows that the proposed approach is very effective for detection of bots in any play styles regardless of feature sets.

# VI. Conclusion

In this paper, we proposed a generic bot detection methodology. In order to detect game bots with different play styles, we examined three aspects, battle, collect and move, of player behaviors which are common in MMORPGs and effective for game bot detection in a low resolution data. Then, players were grouped by their behavioral similarities. Based on the player groups, the proposed method developed a customized local model of each group for bot detection. Since local models were optimized to detect game bots in each player group, the combination of those models could improve the overall performance.

For comparison, the experiment was performed with the data from a game currently in service. Through the experiment, it was verified that the proposed feature sets were effective and the proposed local model approach produced more accurate and stable results for all the play styles.

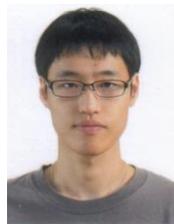

**Yeounoh Chung** received his BS in Electrical and Computer Engineering and his ME in Computer Science from Cornell University, Ithaca, USA, in 2008 and 2009, respectively. He is currently a researcher at Sungkyunkwan University, Suwon, Rep. of Korea. His current research interests include statistical machine learning and data mining

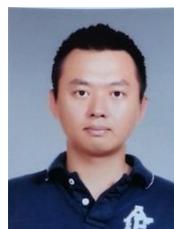

**Chang-yong Park** received his BS in Computer Engineering from Dongguk University, Gyeongju, Rep. of Korea, in 2010. He is currently pursuing his MS in Computer Engineering at Sungkyunkwan University, Suwon, Rep. of Korea. His research interests include software engineering, context-aware recommender system, and latent variable modeling.

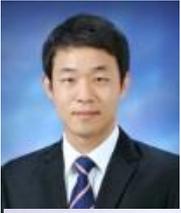

**Noo-ri Kim** received his BS in Computer Engineering from Sungkyunkwan University, Suwon, Rep. of Korea, in 2013. He is currently pursuing his MS in Computer Engineering at Sungkyunkwan University. His research interests include software engineering and latent variable modeling.

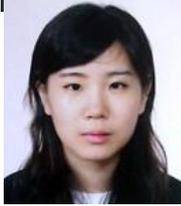

**Hana Cho** received her BS in Computer Engineering from Sungkyunkwan University, Suwon, Rep. of Korea, in 2013. She is currently pursuing her MS in Computer Engineering at Sungkyunkwan University. Her research interests include data mining and artificial intelligence.

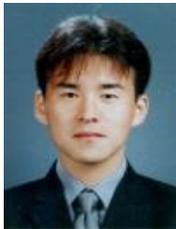

**Taebok Yoon** received his BS in Computer Science from Kongju National University, Kongju, Rep. of Korea, in 2001. He received his MS and PhD in Computer Engineering from Sungkyunkwan University in 2005 and 2010, respectively. He is now an assistant professor at Seoil University, Seoul, Rep. of Korea. His research interests include user modeling, intelligent system, and game artificial intelligence.

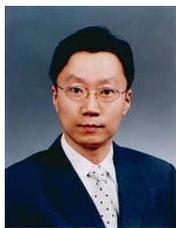

**Hunjoo Lee** received his BS, MS, and PhD in computer science & engineering from Chung-Ang University, Seoul, Rep. of Korea in 1991, 1993, and 1998, respectively. In 1998, he joined ETRI, Daejeon, Rep. of Korea. He was a postdoctoral researcher at Iowa State University, Iowa, USA, from 2001 to 2002. His current research interests include game artificial intelligence, serious game, and smart content.

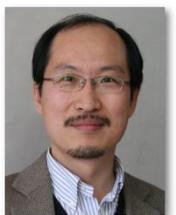

**Jee-Hyong Lee** received his BS, MS, and PhD in Computer Science from the Korea Advanced Institute of Science and Technology (KAIST), Daejeon, Rep. of Korea, in 1993, 1995, and 1999, respectively. From 2000 to 2002, he was an international fellow at SRI International, USA. He joined Sungkyunkwan University, Suwon, Rep. of Korea, as a faculty member in 2002. His research interests include fuzzy theory and application, intelligent system, and machine learning.